\documentclass[preprint,12pt]{elsarticle}

\usepackage{amssymb}
\usepackage{amsmath}
\usepackage{natbib}
\usepackage{graphicx}
\usepackage{blindtext}
\usepackage{cleveref}
\usepackage{multirow}
\usepackage{CJKutf8}
\usepackage{graphicx}
\usepackage{color}
\usepackage{indentfirst} 
\usepackage{subcaption}
\usepackage{natbib}
\usepackage{cite}
\usepackage{cleveref}
\usepackage{multirow}
\usepackage{lineno}
\usepackage{booktabs} 
\usepackage{threeparttable} 
\crefname{figure}{Fig.}{Figs.}
\crefname{equation}{Equation}{Equations}
\crefname{section}{Section}{Sections}
\crefname{subsection}{Subsection}{Subsections}
\crefname{table}{Table}{Tables}

\definecolor{MsBlue}{RGB}{0,112,192} 
\usepackage{lineno}

\journal{Applied Soft Computing}

\begin{document}

\begin{frontmatter}



\title{Dynamic Trend Fusion Module for Traffic Flow Prediction} 
\author[af1]{Jing Chen} 
\ead{cj@hdu.edu.cn}
\author[af2]{Haocheng Ye} 
\ead{yehaocheng@hdu.edu.cn}
\author[af1]{Zhian Ying} 
\ead{yzhian666@gmail.com}
\author[af2]{Yuntao Sun} 
\ead{syt990422@163.com}
\author[af3]{Wenqiang Xu\corref{cor1}} 
\ead{wenqiangxu@cjlu.edu.cn}
\cortext[cor1]{Corresponding Author}
\affiliation[af1]{organization={School of Computer Science and Technology},
            addressline={Hangzhou Dianzi University}, 
            city={Hangzhou},
            postcode={310018}, 
            country={China}}

\affiliation[af2]{organization={ITMO Joint Institute},
            addressline={Hangzhou Dianzi University}, 
            city={Hangzhou},
            postcode={310018}, 
            country={China}}

\affiliation[af3]{organization={College of Economics and Management},
            addressline={China Jiliang University}, 
            city={Hangzhou},
            postcode={314423}, 
            country={China}}
\begin{abstract}
Accurate traffic flow prediction is essential for applications like transport logistics but remains challenging due to complex spatio-temporal correlations and non-linear traffic patterns. Existing methods often model spatial and temporal dependencies separately, failing to effectively fuse them. To overcome this limitation, the \textbf{D}ynamic \textbf{S}patial-\textbf{T}emporal \textbf{T}rend Trans\textbf{former} (\textbf{DST$^2$former}) is proposed to capture spatio-temporal correlations through adaptive embedding and to fuse dynamic and static information for learning multi-view dynamic features of traffic networks. The approach employs the \textbf{D}ynamic \textbf{T}rend \textbf{R}epresentation Trans\textbf{former} (\textbf{DTRformer}) to generate dynamic trends using encoders for both temporal and spatial dimensions, fused via Cross Spatial-Temporal Attention. Predefined graphs are compressed into a representation graph to extract static attributes and reduce redundancy. Experiments on four real-world traffic datasets demonstrate that our framework achieves state-of-the-art performance.

\end{abstract}



\begin{keyword}
Traffic flow prediction \sep Global correlation \sep Spatial–temporal transformer \sep Graph structure
\end{keyword}

\end{frontmatter}

\begin{CJK*}{UTF8}{gbsn}
\section{Introduction}
Traffic prediction is crucial for optimizing road networks, managing congestion, and improving traffic efficiency—key components of urban quality of life and economic vitality \citep{Wen2021, Guo2022}. Accurate predictions enable real-time traffic signal adjustments, dynamic route planning, and congestion pricing, thereby improving flow and reducing delays.

Traditional methods like historical averages \citep{Cui2021}, time series models \citep{LSTM}, and statistical methods \citep{ARIMA} struggle to capture complex non-linear and dynamic patterns in traffic data. Early AI approaches using shallow machine learning models (SVM \citep{SVM}, ANN \citep{ANN}) captured non-linear relationships but faced limitations with high-dimensional spatio-temporal data and generalization. 

\begin{figure}[h]
\centering
\includegraphics[width=0.8\textwidth]{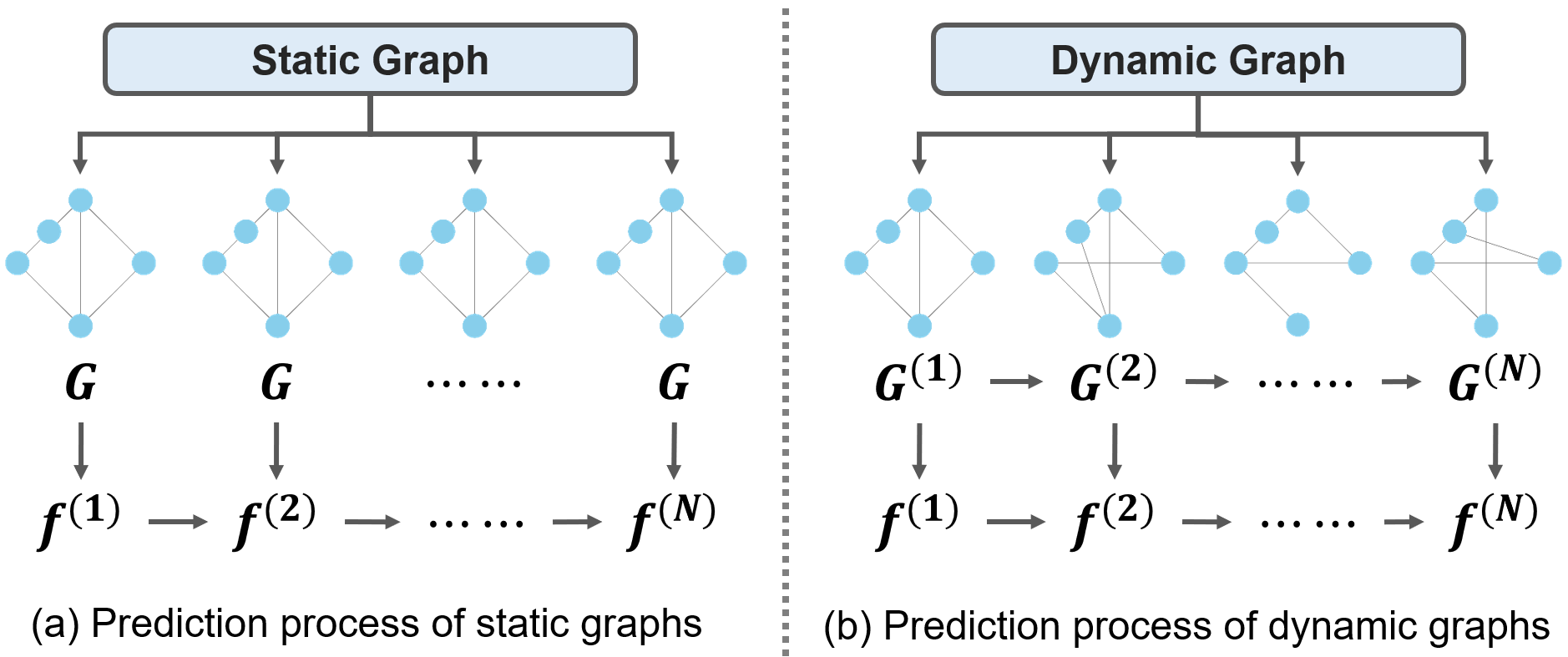}
\caption{The process of generating predictions with static and dynamic graphs}
\label{Fig1}
\end{figure}

Recent deep learning models combine Graph Neural Networks (GNNs) \citep{kipf2017semisupervised} for spatial modeling with Recurrent Neural Networks (RNNs) \citep{li2018} or Transformers \citep{Zheng2020} for temporal modeling, showing promising results.
In temporal modeling, RNNs\citep{RNN}, particularly Long Short-Term Memory (LSTM) networks \citep{LSTM} and Gated Recurrent Units (GRU)\citep{GRU}, effectively capture long-term temporal correlations. Transformers leverage attention mechanisms to capture long-range dependencies.
For spatial modeling, Convolutional Neural Networks (CNNs) capture local spatial dependencies by treating traffic data as images or grids \citep{Yao2018}, but struggle with the non-Euclidean nature of traffic networks. GNNs \citep{kipf2017semisupervised} have become mainstream for spatial modeling, capturing adjacency relationships between nodes, such as with Graph Convolutional Networks (GCNs)\citep{Shao2022, Park2020, Feng2022}.
Current spatio-temporal prediction approaches focus on:

In spatio-temporal models, the construction methods of GNN graph networks can be divided into static graphs and dynamic graphs, as shown in \cref{Fig1}. Static graphs use fixed adjacency matrices based on node distances \citep{li2018} but fail to reflect dynamic traffic flow distributions \citep{GWNET}. Dynamic graph methods generate adaptive adjacency matrices using temporal information \citep{GWNET, Li2022, Zhang2020, D2}, improving performance but increasing computational cost due to per-time-step generation.
    
\begin{figure}[h]
\centering
\includegraphics[width=0.6\textwidth]{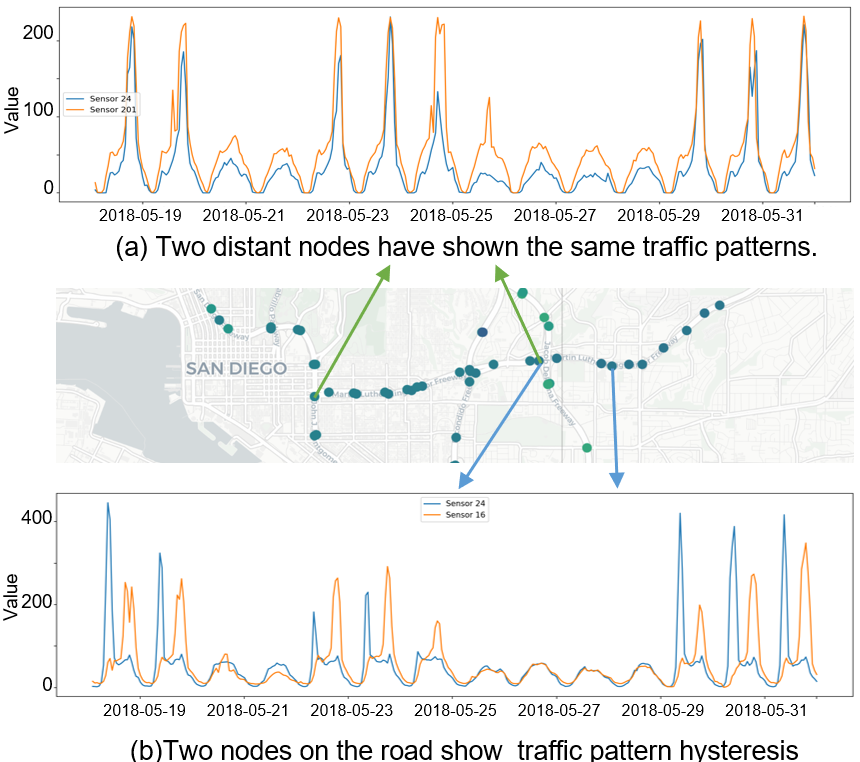}
\caption{Traffic Patterns of Two Groups of Nodes.}
\label{Fig2}
\end{figure}

In terms of spatio-temporal dependencies, classic Transformers are less effective in traffic flow prediction compared to natural language processing. Previous models either ignored unique node patterns \citep{Zheng2020} or lacked interaction between spatial and temporal aspects \citep{Jiang2023}. As shown in \cref{Fig2}, distant nodes can exhibit similar traffic patterns, while nearby nodes may display lagged patterns. Cross-attention mechanisms can help coordinate spatial and temporal relationships.

Based on the limitations of existing models mentioned above, the following constraints can be identified:
\begin{itemize}
    \item Dynamic graph methods require the independent generation of adaptive adjacency matrices at each time step, leading to significant increases in computational resources and time consumption. This restricts the application of models in real-time data processing and long-sequence prediction.
    \item Existing methods have not fully realized the interaction between temporal information and spatial information, failing to effectively capture complex spatio-temporal correlation characteristics, such as the lag in traffic patterns and the similarity of distant nodes. This results in insufficient learning of multi-perspective dynamic information within the traffic network.
\end{itemize}

\begin{figure}[h]
\centering
\includegraphics[width=0.7\textwidth]{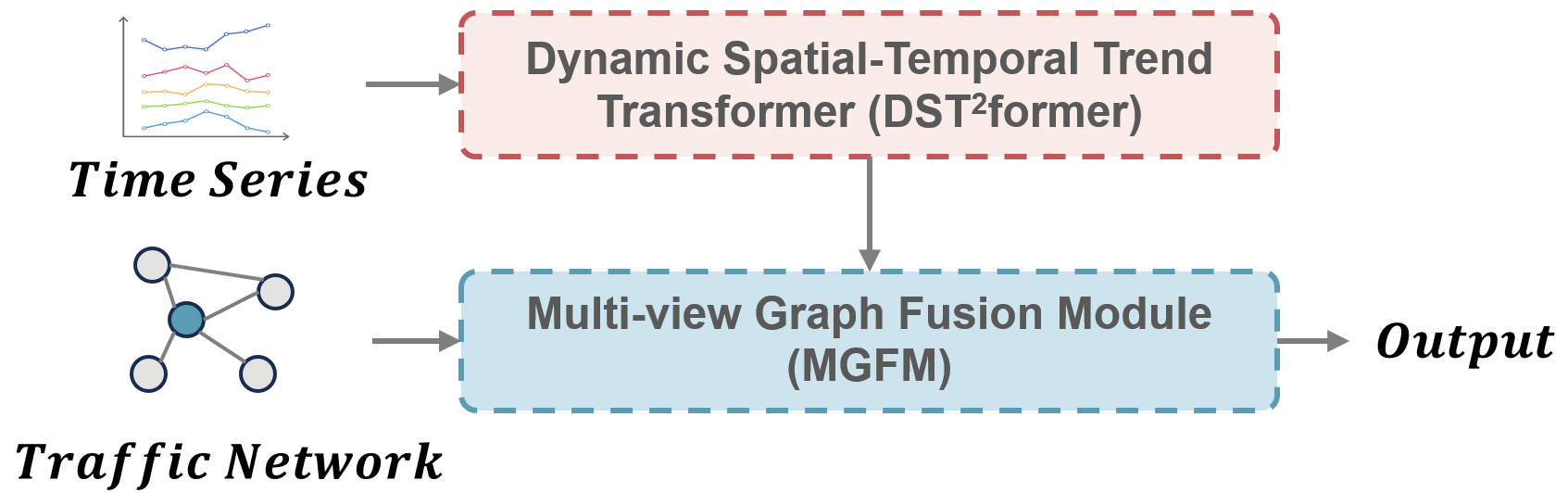}
\caption{Our proposed framwork.}
\label{OverviewofModel}
\end{figure}

To address the above problems, as shown in \cref{OverviewofModel}, the \textbf{D}ynamic \textbf{T}rend \textbf{R}epresentation Trans\textbf{forme}r (DTRformer) is proposed. The \textbf{Dynamic Spatial-Temporal Trend Transformer} is designed to extract dynamic information from temporal data, while the \textbf{Multi-view Graph Fusion Module} captures information from the static road network structure. The main contributions are as follows:

\begin{itemize} 
    \item The traditional dynamic graph generation approach is abandoned, and a novel multi-view graph integration method is proposed to fuse static distance and dynamic trend features. This method reduces the redundancy of static graphs through linear projection and aggregates them with dynamic trends. This integration enables the model to autonomously learn spatial features and capture connectivity at a specific point in time. Additionally, considering the large difference in feature density between the adjacency matrix and dynamic information, the Augmented Residual Attention module is designed to boost information density and enhance the fusion of both types of features. 
    \item A dynamic trend feature generation module based on Transformer is proposed to construct fine-grained multi-step spatio-temporal trends by leveraging its powerful contextual semantic modeling capability. Notably, the cross-attention mechanism is utilized to fuse global temporal and spatial information, facilitating the capture of the evolution of nodes between different time steps by continuously generating their corresponding dynamic trends. 
    \item Extensive experiments conducted on four publicly available datasets demonstrate the effectiveness of the proposed model in the traffic prediction task. Compared to the 18 baselines in the benchmark, the model significantly reduces prediction error and achieves state-of-the-art prediction accuracy. 
\end{itemize}

The remainder of this paper is organized as follows. The next section covers a review of the existing traffic prediction methods. Then, the following two sections define the objectives of the prediction and describe the proposed prediction framework, including details of the specific components. Afterward, the section shows the descriptions of the experimental settings and the results of the proposed method. In the final section, the paper summarizes the key findings and contributions of the proposed model.

\section{Related Works}
Traffic forecasting has been widely studied and applied across various fields. Traditional statistical models such as HI \citep{Cui2021}, ARIMA \citep{Williams2003} and VAR \citep{Chen2001} are commonly employed for time series modeling. Subsequently, machine learning algorithms such as SVM \citep{Jeong2013} and KNN \citep{VanLint2012} were utilized for traffic forecasting. With the advent of deep learning, the performance of time series prediction significantly improved, prompting researchers to integrate spatial and temporal domains to handle more complex data. For instance, \citep{Zhang2021} proposed the ST-ResNet model using residual convolution to predict crowd traffic, which \citep{Yao2018} utilized CNN in the spatial domain and LSTM in the temporal domain. However, these models typically rely on gridded traffic data and overlook non-Euclidean dependencies between nodes.

Recognizing the importance of spatio-temporal modeling, researchers introduced spatio-temporal graph neural networks (STGNN) to capture the complex spatio-temporal correlations in traffic data \citep{Zhang2023b,Li2022,Zhang2018,Seo2018}. Most spatio-temporal graph networks are categorized into two types based on their treatment of the temporal dimension: RNN-based methods \citep{Bai2020,Jang2023,Deng2021} and CNN-based methods\citep{Zhang_Zheng_Qi_2017}. While these methods efficiently decompose spatio-temporal dependencies into spatio-temporal domains, they often requires stacking multiple layers to broaden the field of view.

In recent research, there has been an emergence of models employing the attention mechanism \citep{Jiang2023,Zheng2020,Ye2022,Dwivedi2021} to address aforementioned challenges, as the attention mechanism theoretically offers an infinite field of view and can capture more detailed information. For instance, GMAN \citep{Zheng2020} separately feeds spatial and temporal information into the attention layer, ultimately using a gating mechanism to fuse node information and temporal information. However, its embedding method overlooks the unique traffic patterns of each node. Furthermore, there have been attempts to apply Transformer in spatio-temporal graphs. GT \citep{Dwivedi2021} extended the self-attention mechanism of the Transformer network to graph-structured data, incorporating information transfer and aggregation at the node level. This demonstrated the feasibility of the Transformer structure for information extraction in graph structures, but did not further explore temporal information. Similarly, PDFormer \citep{Jiang2023} modeled the graph using different graph masking methods and addressed the time delays in spatial information propagation to achieve performance enhancements. This approach considers the global and local spatial associations, as well as the temporal relationships between nodes, but unfortunately does not facilitate the interaction between space and time.

Overall, all these approaches limit spatial domain studies to predefined graphs and are not sufficiently capable of modeling the context of dynamic spatio-temporal information. They also cannot directly simulate the inter-temporal effects as well as the dynamic changes on fine granularity. These limitations are addressed by our model through the joint execution of spatio-temporal attention to generate the fusion of dynamic trends and static predefined graphs. 

Meanwhile, as the complexity of the model continues to increase, researchers have turned their attention to embedding approaches that are more compatible with spatio-temporal modeling. As proposed by \citep{Shao2022, Liu2023}, the Identity Embedding method was introduced to adaptively capture spatio-temporal information. This method abandons the graph structure within the spatio-temporal model and simply indexes the spatial sensor nodes before entering linear projection. It is capable of learning the similarity between spatial nodes through self-learning, demonstrating significant predictive performance. However, its effectiveness is limited in complex traffic scenarios due to the lack of information from real road network structures.These embedding approaches inspire the generation part of the dynamic trend. As shown in \cref{OverviewofBaselines}, 14 representative models are selected for classification.

\begin{table}[h]
    \centering
    \caption{Overview of Baselines}
    \begin{threeparttable}
        \begin{tabular}{|c|c|c|}
            \hline
            \textbf{Methods} &\textbf{Category} & \textbf{Component} \\ \hline
            HI & \multirow{3}{*}{Basic Neural Networks} & Historical Dependence \\ \cline{1-1} \cline{3-3}
            LSTM &  & RNN based \\ \cline{1-1} \cline{3-3}
            DLinear &  & Linear + Decomposition \\ \cline{1-3}
            DCRNN & \multirow{3}{*}{Static Graph Based} & GCN + RNN \\ \cline{1-1} \cline{3-3}
            GMAN & & Attention \\ \cline{1-1} \cline{3-3}
            PDFormer & & Attention \\ \cline{1-1} \cline{2-3}
            AGCRN & \multirow{4}{*}{Dynamic Graph Based} & GCN + RNN \\ \cline{1-1} \cline{3-3}
            GWnet &  & GCN + TCN \\ \cline{1-1} \cline{3-3}
            DGCRN & & GCN + RNN \\ \cline{1-1} \cline{3-3}
            MTGNN & & GCN + TCN \\ \cline{1-1} \cline{3-3}
            D$^2$STGNN & & GCN + Attention + RNN\\ \cline{1-1} \cline{2-3}
            ST-Norm & \multirow{3}{*}{Non Graph Based} & Normalization \\ \cline{1-1} \cline{3-3}
            STID & & Identity Embedding + Linear \\ \cline{1-1} \cline{3-3}
            STAEformer & & Identity Embedding + Attention \\ \cline{1-1} \cline{2-3}
        \end{tabular}
    \end{threeparttable}
    \label{OverviewofBaselines}
\end{table}

\section{Preliminaries}
The traffic network is represented by a weighted graph $G=(V, E, A)$, where $V$ is the nodes (i.e., sensors), $|V|=N$ is the number of nodes, $E$ is the set of corresponding edges, and $A \in \mathbb{R}^{N \times N}$ is an adjacency matrix describing the spatial distances between nodes. At each time step $t$, there is a corresponding hidden feature matrix $X^t \in \mathbb{R}^{N \times D}$. Given $T$ time steps on $N$ sensors of the observed traffic flow $X_i^{(t-T)}: X_i^t$, the goal is to predict the next $T$ time steps on $N$ sensors $X_i^{(t+1)}: X_i^{(t+T)}$, where $X_i^t \in \mathbb{R}^{T \times N \times D}$. The mapping relation is as follows:\begin{equation}
\left[X_i^{(t-T+1)}: X_i^t, G\right] \stackrel{f}{\rightarrow} X_i^{(t+1)}: X_i^{(t+T)}.
\end{equation}

\section{Method}
\subsection{Framework structure}
\begin{figure}[h]
\centering
\includegraphics[width=0.8\textwidth]{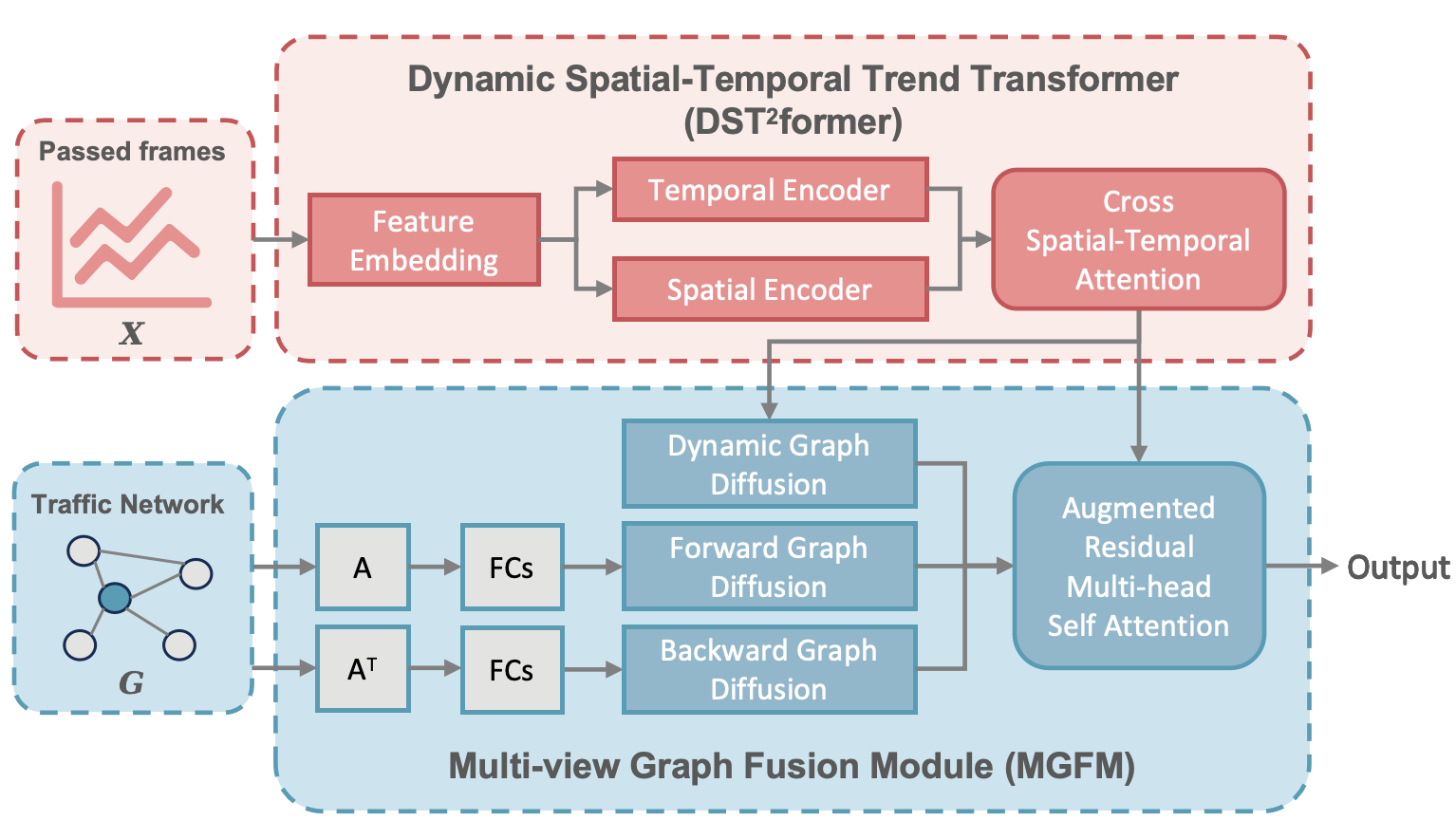}
\caption{General architecture of the proposed DTRformer.}\label{fig1}
\end{figure}
The framework of DTRformer is illustrated in \cref{fig1}. This model accepts two inputs: the adjacency matrix of the traffic network graph and the historical step traffic signals. The original traffic signal first passes through the embedding layer, transforming from the original space  to the hidden space. Subsequently, the information from the hidden space serves as input to the Dynamic Spatial-Temporal Trend Transformer (DST$^2$former), which comprises parallel stacked Temporal Encoder and Spatial Encoder modules. The Cross Spatial-Temporal Attention mechanism is employed to generate spatio-temporal dynamic trend information for $T$-steps. Following this, the trend information is fused with the connectivity information of the predefined graphs, generating a new representation graph via the Multi-view Graph Fusion Module (MGFM). Finally, a simple linear layer processes the output layer to produce the prediction.

\subsection{Dynamic Spatial-Temporal Trend Transformer}
\begin{figure}[h]
\centering
\includegraphics[width=0.55\textwidth]{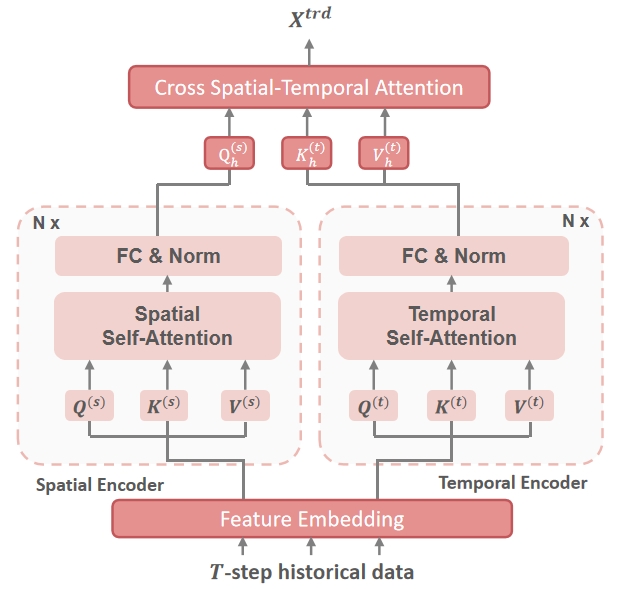}
\caption{Overall structure of the Dynamic Spatial-Temporal Trend Transformer.}
\label{fig2}
\end{figure}
In this section, a spatio-temporal trend learning model is proposed to capture dynamic features. The intensity of traffic flow at two connected nodes varies over time and the flow at a single node affects other nodes. Therefore, modeling dynamically changing traffic flow data is crucial. In natural language processing, the Transformer model attempts to predict the next word based on the context, with complex dependencies between words. Transferred to traffic flow prediction, it is akin to the node-to-node influence in a traffic network and the influence of historical time steps on future time steps.

Specifically, the proposed DST$^2$former is depicted in \cref{fig2}. The output of Embedding Layer denoted as $X$, serves as input. By employing the Self-Attention mechanism, spatial and temporal information in the temporal and spatial dimensions are respectively learned by the adaptive matrix. Finally, Cross Spatial-Temporal Attention is utilized to generate dynamic trend information.

To acquire raw information from known traffic sequences, simple linear mapping is performed in the temporal and spatial dimensions, respectively, to obtain the historical traffic feature embedding $E_{ft}\in \mathbb{R}^{T\times N\times d_f}$ and $E_{fs}\in \mathbb{R}^{T\times N\times d_f}$, where $d_f$ is the dimension of the feature embedding. Considering the periodicity of the real world, a learnable day-of-week embedding dictionary denoted as ${Dict}w \in \mathbb{R}^{N{w \times d_f}}$, and a timestamp-of-day embedding dictionary denoted as ${Dict}d \in \mathbb{R}^{N{d \times d_f}}$ are utilized. Representing $W_t\in \mathbb{R}^T$ as the week-of-day data and $D_t\in \mathbb{R}^T$ as the timestamp-of-day data of the historical $T$ step time series, respectively, these are used as indices to extract the corresponding day-of-week embedding $T_w\in \mathbb{R}^{T\times d_f}$ and timestamp-of-day embedding $T_d\in \mathbb{R}^{T\times d_f}$ from the embedding dictionary. By concatenating and broadcasting them, the embedding of all explicit learnable features $E_f^{\text{exp}} \in \mathbb{R}^{T \times N \times d_e}$ is obtained as follows:
\begin{equation}
E_f^{\text{exp}} = E_{ft} \| E_{fs} \| T_w \| T_d
\end{equation}
where $d_e=4d_f$ and $\|$ denotes the operations of broadcasting and concatenation. Due to the indistinguishability of time and space \citep{Shao2022}, the model adapts to learn temporal and spatial information for each node in an adaptive manner, aiming to capture the unique traffic patterns of nodes. Specifically, by initializing a random matrix $E_a^{\text{imp}} \in \mathbb{R}^{N \times d_a}$, and concatenating it with historical traffic information and temporal identity information, the final output of the Embedding Layer $X \in \mathbb{R}^{T \times N \times D}$ is obtained:
\begin{equation}
X=E_f^{\text{exp}} \| E_a^{\text{imp}}
\end{equation}
where $D=d_e+d_a$.

\subsubsection{Spatial Encoder and Temporal Encoder}
With the parallel encoder module, trends unique to each of the temporal and spatial dimensions can be captured. Due to the field of view advantage, each node can access information from other nodes, and similarly, each time step can access information from other time steps. In the traditional Transformer, multi-head self-attention serves as the key component, enabling the model to capture long-distance dependencies in the input sequence by approximating the mathematical expectation for the entire input sequence divided into different representation spaces. Given $X$ as input, the query, key, and value matrices are obtained through spatial transformer layers as follows:
\begin{equation}
Q^{(s)}=XW_Q^{(s)}, K^{(s)}=XW_K^{(s)}, V^{(s)}=XW_V^{(s)}
\label{eq:MSA_project}
\end{equation}
where $W_Q^{(s)}$ ,$W_K^{(s)}$ and $W_V^{(s)}\in \mathbb{R}^{D \times D}$ are learnable parameters. Then calculation of the self-attention score can be written as:
\begin{equation}
A^{(s)}=\operatorname{softmax}\left(\frac{Q^{(s)} K^{(s)^T}}{\sqrt{D_{Q^{(s)} K^{(s)}}}}\right) V^{(s)}.
\label{eq:MSA_AS}
\end{equation}
where $A^{(s)}$ captures the spatial relations in different time steps. Next, all the tokens are included, and the equation is extended to multi-head self-attention as follows:
\begin{equation}
     H^{(s)}=concat(A_1^{(s)},\ ...,A_h^{(s)})W^O,
     \label{eq:MSA_concat}
\end{equation}
where $h$ is the number of heads set as a hyperparameter, $W^O$ is the output matrix, and $H^{(s)}\in \mathbb{R}^{T\times N\times D}$ refers to a unique spatial node with different connection relations. Similarly, the temporal attention layer performs as follow:
\begin{equation}
    H^{(t)}=SelfAttention\left(A_1^{\left(t\right)},\ ...,A_h^{\left(t\right)}\right),
\end{equation}
where $SelfAttention$ follows Equations \ref{eq:MSA_project},\ref{eq:MSA_AS},\ref{eq:MSA_concat} and $H^{(t)}\in \mathbb{R}^{N\times T\times D}$ captures the temporal relations in different spatial nodes. Notably, residuals and layer normalization are applied.

\subsubsection{Cross Spatial-Temporal Attention}
The spatio-temporal encoder enables us to extract features across spatial and temporal dimensions. However, effectively fusing these features presents a challenge: the information changes in the spatial and temporal dimensions each have their own trends, yet they influence each other. To address this issue, a cross spatio-temporal attention mechanism is introduced to capture their joint evolution.

For a given node $n$, the global nature of the attention mechanism allows it to capture the evolution of trends over the past $T$ time steps, understanding the attention relationships across time. Similarly, at a specific time step $t$, the model can capture the cross-connection relationships between different nodes, enabling it to understand spatial interactions at different time steps. This capability allows the model to comprehensively understand the spatio-temporal connections in the data, providing a more accurate foundation for the analysis and prediction of spatio-temporal data.

To simplify the computation steps, only the information from the adaptive matrix is focused on, and the projection of temporal identity information and the input information $E_f^{\text{exp}}$ is temporarily not discussed. Spatial information is used as the query matrix $Q^{(s)}$, and temporal information is used as the key matrix $K^{(t)}$ and value matrix $V^{(t)}$. By utilizing $Q^{(s)}$ and $K^{(t)}$, the feature weight graph ${G}^{w}{\in \mathbb{R}}^{T\times N\times d_a}$ is obtained, as shown in the following formula:
\begin{equation}
{G}^{w} = softmax\left(\frac{Q^{(s)}_{h}K^{(t)^T}_{h}}{\sqrt{D_{Q^{(s)}_{h}K^{(t)}_{h}}}}\right),
\end{equation}
where $Q^{(s)}_{h}$ and $K^{(t)}_{h}$ are obtained by projecting $H^{(s)}$ and $H^{(t)}$, respectively.

Through this computation, each spatial node can focus on the temporal features across all time steps. In practical traffic scenarios, this is equivalent to querying the current data from all $N$ sensors at each time step $i$. $T$ dynamically changing feature weight graphs, $G_{t-T+1}^w, \ldots, G_t^w {\in \mathbb{R}}^{T\times N\times d_a}$, are obtained, reflecting the correlation between different time steps and spatial information.

Let $G_t^w$ denote the weight graph for the $i$-th time step, and $X_i^{\text{trd}}$ denote the output for the $i$-th time step. The weight graph $G_i^w$ is then multiplied with the corresponding temporal value matrix $V_i^{(t)}$, calculated as follows:
\begin{equation}
X_i^{\text{trd}} = G_i^w V_i^{(t)},
\end{equation}
where $X_i^{\text{trd}} {\in \mathbb{R}}^{1\times N\times d_a}$. Extending this to $T$ time steps, $X_{t-T+1}^{\text{trd}}, \ldots, X_t^{\text{trd}}$ represents the collection of dynamic trends across all time steps.

The final output is represented as ${H}^{(st)} {\in \mathbb{R}}^{T\times N\times D}$, which includes the dynamic trends $X^{\text{trd}}$, as well as historical traffic information and temporal identity information $E_f^{\text{exp}}$. Notably, the output of temporal attention $H^{(t)}$ is used as a residual connection, and for simplicity, some details of the feedforward layers are omitted.

\subsection{Multi-view Graph Fusion Module}
\begin{figure}[h]
\centering
\includegraphics[width=0.45\textwidth]{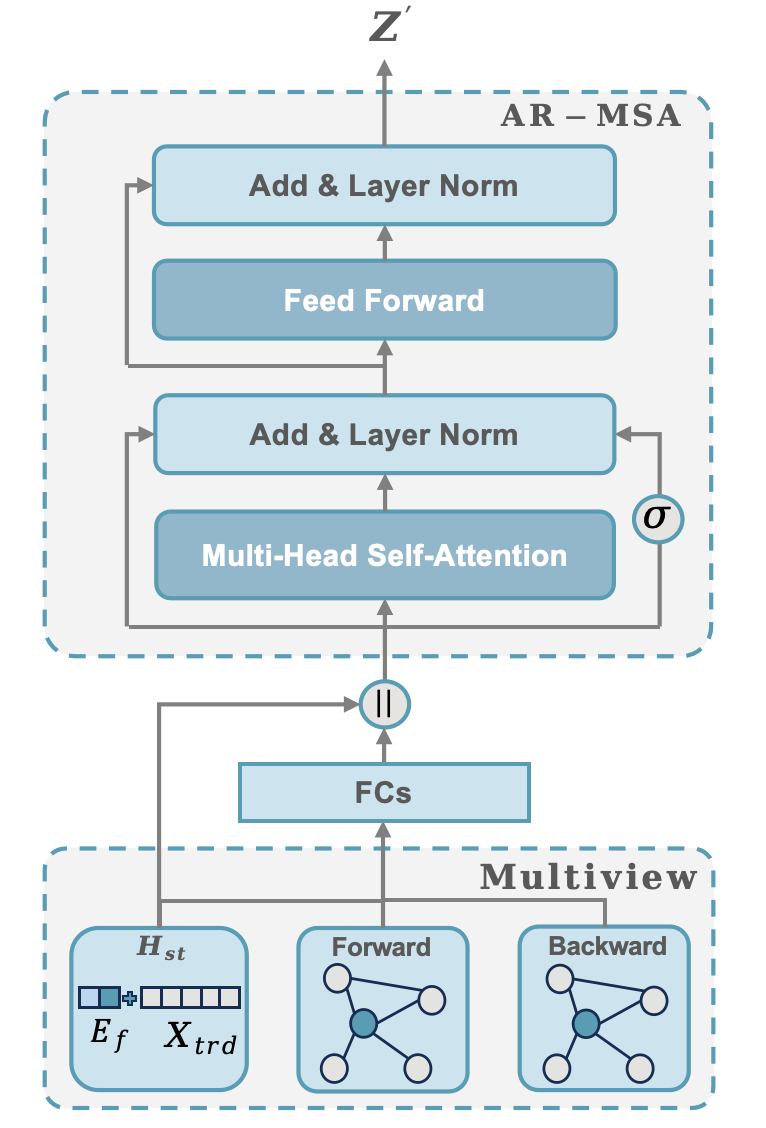}
\caption{Overall structure of the Multi-view Graph Fusion Module.}\label{fig3}
\end{figure}

In traffic flow, the importance of connectivity relationships between nodes changes over time. Therefore, this module aims to explore the explicit connectivity relationships at different time steps. It combines the $H^{(st)}$ spatio-temporal trend information with the connectivity of that node in the static graph to amplify or reduce the influence of that point. The overall structure is shown in \cref{fig3}.

\subsubsection{Dynamic-Static Fusion}
In the road network, considering the direction of diffusion in both directions, this property is utilized to capture static information. By distinguishing between forward and backward transfers, traffic flow and driving behavior in different directions can be modeled more accurately. Firstly, the adjacency matrix $A$ is leveraged for static information capture. The bi-directional transfer relationship of the road network is represented as forward transfer, which indicates the degree of influence of a spatial node on other nodes (i.e., the out-degree of a spatial node), normalized as forward transfer: $A^f=\frac{A}{rowsum(A)}$. Similarly, reverse transfer represents the intensity of influence of a spatial node on other nodes (i.e., the out-degree of the node), normalized as: $A^b=\frac{A^T}{rowsum\left(A^T\right)}$. To reduce sparsity in the original adjacency matrix, $A^f$ and $A^b$ are normalized and fed into the linear layer for dimensionality reduction. $A^f$ and $A^b$ are transformed into $G^f\in \mathbb{R}^{N\times d_n}$ and $G^b\in \mathbb{R}^{N\times d_n}$, respectively:
\begin{equation}
    G^f=\sigma(A^fW)
\end{equation}
where $\sigma$ represents the $ReLU$ activation function to remove weak connections and $W$ is the learnable parameter of the linear layer. The computational for $G^b$ the same.
Subsequently, the connection is made via the broadcast method:
\begin{equation}
    X^{\text{hid}} =\ G^f||G^b||X^{\text{trd}} ,
\end{equation}
where $X^{\text{hid}} \in \mathbb{R}^{N\times(d_a+2\times d_n)}$. Then, the MLP layer with residual connections is used to encode the information to fuse the spatial information from multiple viewpoints. The first $l$ layer can be represented as:
\begin{equation}
    {(X^{\text{hid}})}^{l+1}={FC}_2^l(\sigma({FC}_1^l{(X^{\text{hid}})}^l))+{(X^{\text{hid}})}^l,
\end{equation}
where, the final output $({X^{\text{hid}})}^L$ is taken, and after dimensionality reduction through the linear layer, an output ${X^{\text{raw}}}$ matching the size of ${X^{\text{trd}}}$ is obtained. At this stage, as $E_f^{\text{exp}}$ and $X^{\text{trd}}$ are combined, $X^{\text{raw}}$ captures the complete connectivity relationships, representing a preliminary fusion at the $i$-th time step.

\subsubsection{Augmented Residual Multi-head Self Attention}
This section aims to design a mechanism to adjust the temporal identity information by aligning the feature weights $X^{\text{raw}}$ of both the static graph information and the dynamic trend in the middle for the final prediction. Firstly, $E_f^{\text{exp}}$ is combined with $X^{\text{raw}}$ by concatenating them to obtain the input data $Z$. Subsequently, these input data are passed into the attention layer with enhanced residual connectivity. The model dynamically adjusts the weights of features with specific temporal identity information. The reinforced residuals are designed to maximize the information density to mitigate the effects of sparsity due to the static graph structure.
Given that the inputs to the specific defined first $l$ The input to the layer is $Z^l$ ,which undergoes an activation function to obtain augmented residual ${\bar{Z}}^l$:
\begin{equation}
    {\bar{Z}}^l=\sigma(Z^l),
\end{equation}
where $Z^l$ , and ${\bar{Z}}^l\in \mathbb{R}^{N\times D}$, and $\sigma$ represents the $GeLU$ activation function. Finally, the final output $Z^{l+1}$ is obtained after layer normalization. The calculation process is as follows:
\begin{equation}
    Z^{l+1}=LN(Z^l+{\bar{Z}}^l+H^l),
\end{equation}
where $H^l$ is the output of the multi-head self-attention, computed in accordance with Equations \ref{eq:MSA_project},\ref{eq:MSA_AS},\ref{eq:MSA_concat}.
Finally, the final output of the global spatio-temporal encoder is obtained after a feed-forward neural network with residual connectivity and layer normalisation, denoted as $Z^\prime\in \mathbb{R}^{T\times N\times D}$ .

\subsection{Output and Loss}
To synchronize the prediction results for the next $P$ time steps, a linear transformation is directly applied to the outputs of the Multi-view Graph Fusion Module ${\ Z}^\prime$ after AR-MSA to generate the final prediction, denoted as $\check{Y} \in \mathbb{R}^{T\times N \times C_{out}}$. Given the ground truth value $Y \in \mathbb{R}^{T \times N \times C_{out}}$, the model is optimized using the Mean Absolute Error (MAE) loss:
\begin{equation}
    L(\check{Y},Y;\Theta) = \frac{1}{TNP}\left(x+a\right)^n=\sum_{i=1}^{P}\sum_{j=1}^{N}\sum_{k=1}^{C_{out}}\left|{\check{Y}}_{ijk}-Y_{ijk}\right|,
\end{equation}
where $P$ is the number of predicted steps, $N$ is the number of nodes, and $C_{out}$ is the dimension of the output.

\section{Experiments}
In this section, a comparison between DTRformer and other baselines is presented using four spatio-temporal traffic datasets. The experimental settings, including datasets, baselines, and parameter settings, are first introduced. Then, experiments are conducted to compare the performance of DTRformer with other baselines. Additionally, ablation experiments are designed to evaluate the impact of baseline architectures, components, and training strategies.

\subsection{Datasets}
Experiments are conducted on four commonly used publicly available large-scale datasets, all of which contain tens of thousands of timesteps and hundreds of sensors. PEMS03, PEMS04, PEMS07, and PEMS08 are published by \citet{Guo2019}. Each of these four datasets was constructed from four regions in California. 
\begin{table}[h]
    \centering
    \caption{Statistics of Datasets}
    \begin{tabular}{c|c|c|c}
    \hline\hline
         Datasets & Nodes & Samples & Sample Rate  \\   \hline
         PEMS03 & 358 & 26208 & 5min  \\    \hline
         PEMS04 & 307 & 16992 & 5min  \\    \hline
         PEMS07 & 883 & 28224 & 5min  \\    \hline
         PEMS08 & 170 & 17856 & 5min  \\    \hline\hline
    \end{tabular}
    \label{datasets}
\end{table}
All these data were collected from Caltrans Performance Measurement System (PEMS), and the spatial adjacency matrix of each dataset was constructed using the actual distance-based road network. The statistical information is shown in \cref{datasets}.

The data is normalized using the Z-Score method. Approximately 60\% of the data is allocated for training, 20\% for validation, and the remaining 20\% for testing. Additionally, the weighted neighbor matrix is constructed in the same way as DCRNN \citep{li2018}. Considering each traffic sensor as a vertex, it can be represented as follows:

\begin{equation}
    A_{v_i,v_j} = 
    \begin{cases}
        exp \left(-\frac{d_{v_i,v_j}^2}{\sigma^2}\right), & \text{if } exp \left(-\frac{d_{v_i,v_j}^2}{\sigma^2}\right) \geq \theta \\
        0, & \text{otherwise } 
    \end{cases}
    ,
\end{equation}
where $d_{v_i,v_j}^2$ denotes the distance from node $v_i$ to node $v_j$ of the road network distance, $\sigma$ is the standard deviation, and $\theta$ (assigned to 0.1) denotes the threshold value.

\subsection{Baselines}
In this study, several widely used baselines in the field are selected for comparison, and the results are presented below:
\begin{itemize}
    \item HI\citep{Cui2021}: Historical Inertia model, which takes advantage of inertia and trends in historical data, using past data as a baseline to predict future time series.
    \item DLinear\citep{Zeng_Chen_Zhang_Xu_2023}: DLinear decomposes raw data into a trend component and a remainder (seasonal) component using a moving average kernel. Then, two one-layer linear layers are applied to each component, and the final prediction is obtained by summing them up.
    \item VAR\citep{VAR_SVR}: Vector Auto-Regression (VAR) can be used for time series forecasting.
    \item SVR\citep{VAR_SVR}: : Support Vector Regression (SVR) is another classical time series analysis model which uses linear support vector machine for the regression task.
    \item ARIMA\citep{ARIMA_EXP}: This is a traditional and widely used method in time series prediction, which integrates autoregression with moving average model.
    \item STGODE\citep{STGODE}: Spatial-Temporal Graph Ordinary Differential Equation Networks is a model that combines tensor-based ordinary differential equations with graph neural networks to capture spatial-temporal dependencies.
    \item LSTM\citep{LSTM}: Long Short-Term Memory is a specialized type of recurrent neural network designed to effectively capture long-term dependencies in sequential data, overcoming the vanishing gradient problem faced by traditional RNNs in learning from long sequences.
    \item STID\citep{Shao2022}:STID explores and proposes spatial embedding and temporal periodicity embedding approach and captures spatio-temporal patterns based on MLP.
    \item Graph WaveNet\citep{GWNET}: Graph WaveNet introduces a adaptive dependency matrix that can be learned through node embedding. Gated TCNs and GCNs are stacked layer by layer to jointly capture spatial and temporal dependencies.
    \item DCRNN\citep{li2018}: Diffusion Convolutional Recurrent Neural Network proposes diffusion convolutional layers and incorporates both spatial and temporal dependencies into a sequence-to- sequence framework for traffic flow prediction.
    \item AGCRN\citep{Bai2020}: AGCRN comes out with an adaptive graph convolution recurrent network model, which incorporates the features of graph convolution networks and recurrent neural networks by adaptively adjusting the graph convolution operation.
    \item MTGNN\citep{wu2020}: MTGNN effectively captures complex relationships in time series data by building correlation graphs between variables and applying graph convolution module.
    \item DGCRN\citep{Li2022}: DGCRN models dynamic graphs and designs a novel dynamic graph convolutional loop module to capture spatio-temporal patterns in seq2seq architectures.
    \item D$^2$STGNN\citep{D2}: D$^2$STGNN decouples the diffusion signal and inherent signal from traffic data. A spatio-temporal localized convolution is designed to model the hidden diffusion time series. The RNN and self-attention mechanism are jointly used to model the hidden inherent time series. Furthermore, the dynamic graph learning module is adopted.
    \item STNorm\citep{Deng2021}: STNorm uses temporal normalisation and spatial normalisation to refine high frequency components and local components respectively, for distinguishing dynamics in spatial perspective and dynamics in the temporal perspective.
    \item GMAN\citep{Zheng2020}: An encoder-decoder structure based on stacked attention modules with a gating mechanism to fuse spatio-temporal factors on traffic information.
    \item PDFormer\citep{Jiang2023}: PDFormer adopts a dynamic long-range Transformer structure and proposes a spatial attention module.
    \item STAEformer\citep{Liu2023}: STAEformer proposes a novel embedding approach and captures spatio-temporal patterns based on an attention mechanism.
\end{itemize}

\subsection{Experimental Settings}
Our experimental setup follows existing methods to ensure fairness in comparison. For PEMS03, PEMS04, and PEMS08, the data split ratio for training, validation, and testing was set to 6:2:2. The proposed model was implemented using PyTorch 1.9.1 on an NVIDIA 4090 GPU. Consistent with previous studies, the model was trained using the Adam optimizer with an initial learning rate of 0.001. The embedding dimension in the model, $d_f$, is set to 24, the dimension of the pre-defined graph $d_a$ is 100, and the number of heads is 4. Hyperparameters tuned in the model include the number of encoder blocks $N$. The historical time step $T$ is set to 12 (1 hour) to predict future traffic conditions over $T'$ = 12 time steps. If the validation loss does not converge within 10 consecutive steps, the early stop mechanism is applied. The code is available at: (https://github.com/HITPLZ/DSTRformer). Three widely used metrics are adopted to evaluate traffic prediction performance: Mean Absolute Error (MAE), Root Mean Square Error (RMSE), and Mean Absolute Percentage Error (MAPE). The formulas are as follows:
\begin{equation}
\operatorname{MAE}(\theta, \bar{\theta})=\frac{1}{\mathrm{~T}} \sum_{\mathrm{i} \in \mathrm{T}}\left|\theta_{\mathrm{i}}-\bar{\theta}_{\mathrm{i}}\right|
\end{equation}
\begin{equation}
\operatorname{RMSE}(\theta, \bar{\theta})=\frac{1}{\mathrm{~T}} \sqrt{\sum_{\mathrm{i} \in \mathrm{T}}\left|\theta_{\mathrm{i}}-\bar{\theta}_{\mathrm{i}}\right|}
\end{equation}
\begin{equation}
\operatorname{MAPE}(\theta, \bar{\theta})=\frac{1}{\mathrm{~T}} \sum_{\mathrm{i} \in \mathrm{T}} \frac{\left|\theta_{\mathrm{i}}-\bar{\theta}_{\mathrm{i}}\right|}{\theta_{\mathrm{i}}}
\end{equation}
where $\theta_i$ denotes the $i$-th ground truth, $\bar{\theta}_{\mathrm{i}}$ represents the $i$-th predicted values, and $T$=12 in our experiments is the indices of observed samples. The model is trained on the training dataset, model selection is performed on the validation dataset, and performance is reported on the test dataset.

\subsection{Experimental Results}
\subsubsection{Error Metrics}
As shown in \cref{table2}, our approach achieves the best results on most of the metrics on the four datasets. HI represents the effectiveness of traditional methods for spatio-temporal prediction of time series, and achieves the worst results due to the lack of spatial modelling. DCRNN and Graph WaveNet are two typical spatio-temporal coupling models that still lead in performance. MTGNN improves on information extraction and proposes the learning potential adjacency matrices for further improvement. DGCRN improves on DCRNN by using a dynamic graph approach and achieves improved performance. GMAN achieves  good results in long-term prediction using an attention-based encoder-decoder architecture. PDFormer focus on short-term traffic prediction and shows improvement based on Transformer. STID and STAEformer explore the embedding approach and achieve impressive results even with simple feature extraction. In short, this table demonstrates the superiority of our DTRformer.

\subsubsection{Performance Evaluation Metrics}
\begin{figure}[h]
\centering
\includegraphics[width=0.68\textwidth]{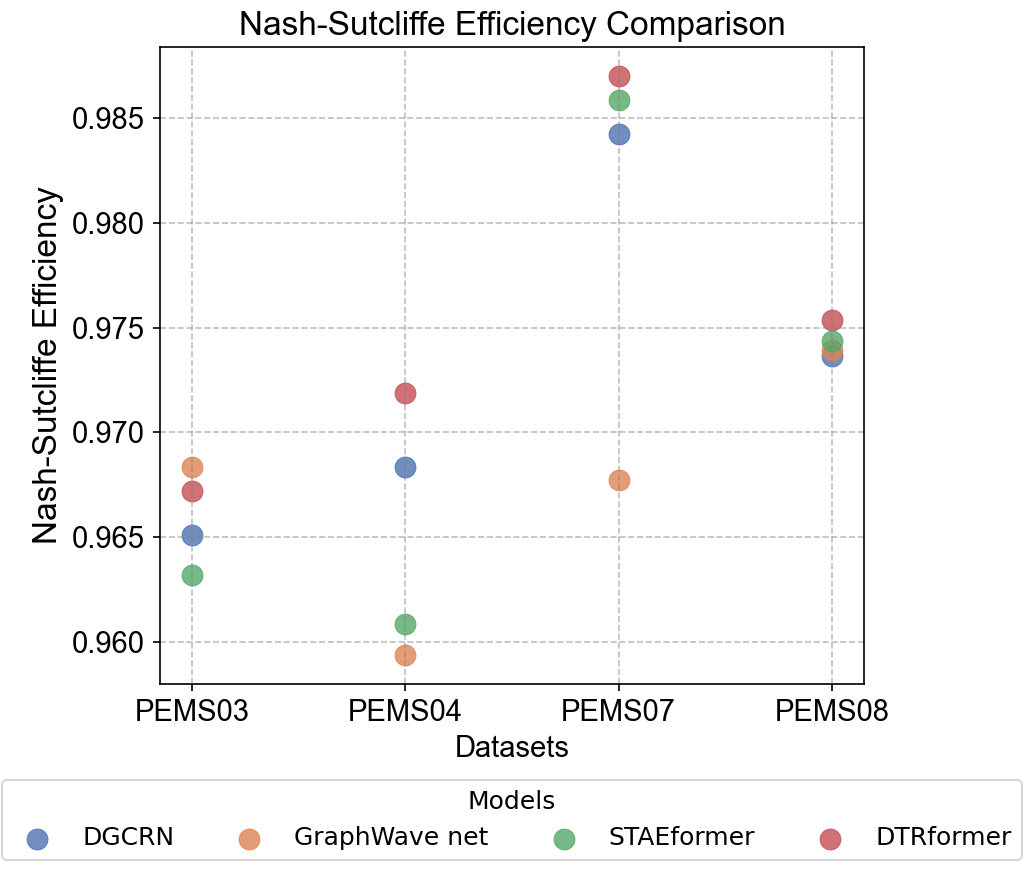}
\caption{Nash-Sutcliffe Efficiency (NSE) Analysis.}
\label{r2}
\end{figure}

Based on the results shown in \cref{r2}, the DTRformer model achieved excellent predictive performance on four different traffic forecasting datasets, with a Nash-Sutcliffe Efficiency (NSE) exceeding 0.96 in all cases, outperforming other comparable models. This highlights the effectiveness of the dynamic spatio-temporal relationship fusion mechanism we proposed, providing an efficient and reliable solution for spatio-temporal data modeling and forecasting.

\begin{figure}[h]
\centering
\includegraphics[width=1\textwidth]{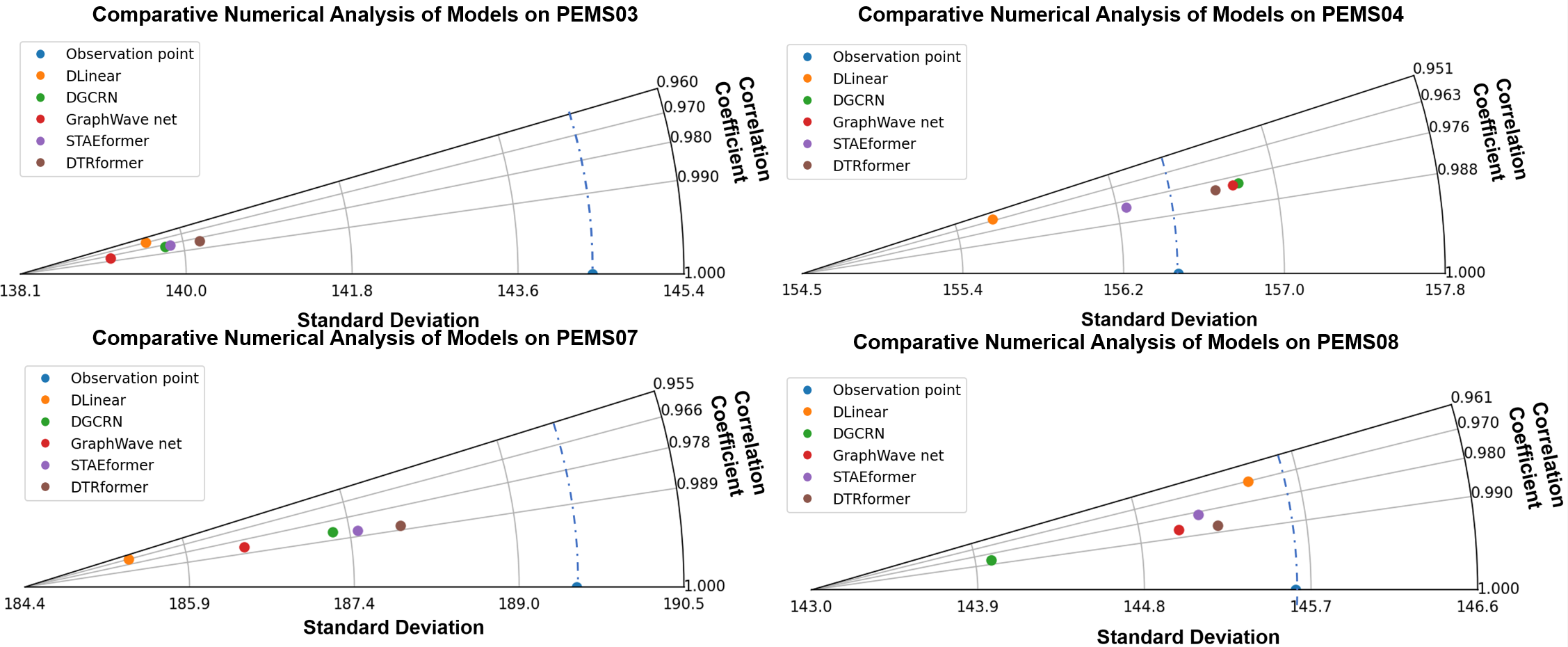}
\caption{Tylor Diagram Analysis}
\label{Tylor}
\end{figure}

Additionally, the Taylor diagram shown in \cref{Tylor} is presented to provide a comprehensive visualization of the model's performance. The angular position reflects the correlation between the model's prediction results and the observed data, while the radial distance represents the standard deviation of the model's simulation results. It can be seen that in all four datasets, DTRformer has the highest correlation coefficient and the standard deviation is closest to the observed data, indicating the best performance.

\begin{figure}[h]
\centering
\includegraphics[width=0.9\textwidth]{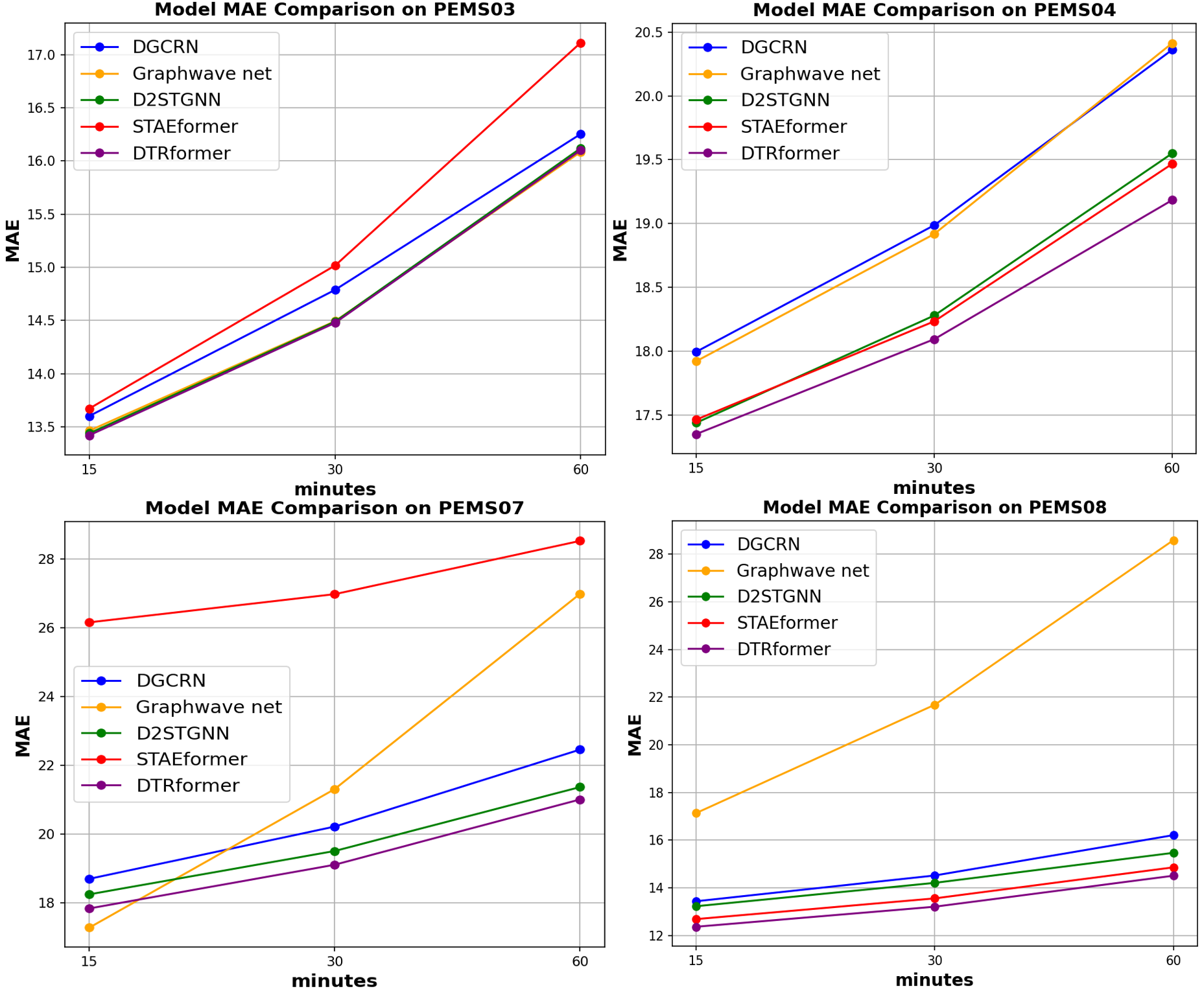}
\caption{Comparison of Model Accuracy at Three Time Points}
\label{Margin}
\end{figure}

Furthermore, as shown in \cref{Margin}, we selected DCRNN, Graphwave Net, D$^2$STGNN, and STAEformer to compare their prediction accuracy with DTRformer at 15-minute, 30-minute, and 60-minute intervals. The results demonstrate that our model achieved the best performance in predictions at almost all time points.

\begin{table}[htb]
\caption{Performance on PEMS03, PEMS04, PEMS07, and PEMS08.}
\resizebox{1\textwidth}{!}{
\begin{tabular}{c|ccc|ccc|ccc|ccc}
\hline\hline
Dataset                       & \multicolumn{3}{c|}{PEMS03} & \multicolumn{3}{c|}{PEMS04}                        & \multicolumn{3}{c|}{PEMS07}                       & \multicolumn{3}{c}{PEMS08}                        \\ \hline\hline
Metric      & MAE            & RMSE           & MAPE                  & MAE            & RMSE           & MAPE             & MAE            & RMSE           & MAPE            & MAE            & RMSE           & MAPE            \\ \hline\hline
HI               & 32.62 & 49.89 & 30.60\%            & 42.35          & 61.66          & 29.92\%          & 49.03          & 71.18          & 22.75\%         & 36.66          & 50.45          & 21.63\%         \\
ARIMA       & 35.31 & 47.59 & 33.78\%            & 33.73          & 48.80          & 24.18\%          & 38.17          & 59.27          & 19.46\%         & 31.09          & 44.32          & 22.73\%         \\
VAR               & 23.65 & 38.26 & 24.51\%            & 23.75          & 36.66          & 18.09\%          & 75.63          & 115.24          & 32.22\%         & 23.46          & 36.33          & 15.42\%         \\
SVR                    & 21.97 & 35.29 & 21.51\%            & 28.70          & 44.56          & 19.20\%          & 32.49          & 50.22          & 14.26\%         & 23.25          & 36.16          & 14.64\%         \\

DLinear                         & 21.35 & 34.48 & 21.52\% & 37.52          & 62.21
& 17.26\%          & 23.23          & 37.77          & 9.64\%          & 43.68         & 65.24          & 20.96\%          \\

LSTM                         & 16.58 & 28.12 & 16.25\% & 22.22          & 35.32
& 15.28\%          & 23.23          & 37.77          & 9.64\%          & 15.23          & 24.17          & 10.21\%          \\
STGODE                         & 16.50 &27.84  &16.69\% & 20.84          & 32.82
& 13.77\%          & 22.29          & 37.54          & 10.14\%          & 16.81        & 25.97          & 10.62\%          \\
STID    &  15.33 & 27.40 & 16.40\%                  & 18.38          & 29.95 & 12.04\%          & 19.61          & 32.79          & 8.30\%          & 14.21          & 23.28          & 9.27\%          \\

GWNet                         & {14.59} & 25.24 & 15.52\% & 18.53          & {29.92}          & 12.89\%          & 20.47          & 33.47          & 8.61\%          & 14.40          & 23.39          & 9.21\%          \\
DCRNN    &  15.54 & 27.18 & 15.62\%                   & 19.63          & 31.26          & 13.59\%          & 21.16          & 34.14          & 9.02\%          & 15.22          & 24.17          & 10.21\%         \\
AGCRN      & 15.24 & 26.65 & 15.89\%                & 19.38          & 31.25          & 13.40\%          & 20.57          & 34.40          & 8.74\%          & 15.32          & 24.41          & 10.03\%         \\
MTGNN   & 14.85 & \textbf{25.23} & 14.55\%                    & 19.17          & 31.70          & 13.37\%          & 20.89          & 34.06          & 9.00\%          & 15.18          & 24.24          & 10.20\%         \\
DGCRN   & 14.60 & 26.20 & 14.87\%                    & 18.84          & 30.48          & 12.92\%          & 20.04          & 32.86          &  8.63\%          & 14.77          & 23.81          &  9.77\%         \\

D$^2$STGNN    &  14.63 & 26.31 & 15.32\%                  & 18.32          & 29.89 & 12.51\%          & 19.49          & 32.59          & 8.09\%          & 14.10         & 23.36          & 9.33\%          \\

STNorm    & 15.32 &25.93 & \textbf{14.37\%}                   & 18.96          & 30.98          & 12.69\%          & 20.50          & 34.66          & 8.75\%          & 15.41          & 24.77          & 9.76\%          \\
GMAN  & 16.87 & 27.92 & 18.23\%                       & 19.14          & 31.60          & 13.19\%          & 20.97          & 34.10          & 9.05\%          & 15.31          & 24.92          & 10.13\%         \\
PDFormer & 14.94 & 25.39 & 15.82\% & 18.36          & 30.03          & 12.00\% & 19.97          & 32.95          & 8.55\%          & 13.58          & 23.41          & 9.05\%          \\

{STAEformer}   & 15.35 & 27.55 & 15.18\%            & {18.22} & 30.18          & \textbf{11.98}\%          & 19.14 & 32.60 & 8.01\%
& {13.46} & {23.25} & {8.88\%} \\ 
\textbf{DTRformer}   & \textbf{14.50} & {25.45} & 14.94\%            & \textbf{18.00} & \textbf{29.58}          & {12.30}\%          & \textbf{18.99} & \textbf{32.23} & \textbf{7.93\%} & \textbf{13.17} & \textbf{22.85} & \textbf{8.66\%} \\ 
\hline\hline
\end{tabular}
}
\label{table2}
\end{table}

\subsection{Ablation Study}
To assess the validity of each component in DTRformer, an ablation study is conducted on six variants of the model.
\begin{table}[h]
\caption{Ablation Study on PEMS03, PEMS04, PEMS07, and PEMS08.}
\resizebox{1\textwidth}{!}{
\begin{tabular}{c|ccc|ccc|ccc}
\hline\hline
Dataset        & \multicolumn{3}{c|}{PEMS03}    & \multicolumn{3}{c|}{PEMS04}   & \multicolumn{3}{c}{PEMS08}                        \\ 
\hline\hline
Metric         & MAE            & RMSE           & MAPE             & MAE            & RMSE           & MAPE            & MAE            & RMSE           & MAPE            \\ 
\hline\hline
w/o $E^{\text{imp}}_a$     &14.71	&25.67	&15.10\%     &18.28	&29.75	&12.57\%          & 16.13          & 26.52          & 11.10\%          \\
w/o $Transformer$      &14.94	&26.46	&15.55\%          &18.56	&30.32	&12.94\%          & 14.07          & 23.17          & 9.18\%          \\ 
\hline
w/o $A^f$    &14.75	&25.60 	&15.16\%          &18.08	&29.77	&12.32\%          & 13.26          & 22.90          & 8.71\%            \\
w/o $A^b$   &14.79	&25.87	&15.13\%          &18.09	&29.77	&12.32\%         & 13.41          & 22.90          & 8.74\%         \\
w/o $A^f  \&  A^b$   &14.96	&26.14	&15.13\%          &18.16	&29.77	&23.39\%         & 13.41          & 23.12          & 8.72\%         \\ 
\hline
w/o $augmented~residual $   &14.73	&25.46	&15.05\%     &18.11	&29.83	&12.36\%         & 13.26          & 22.88          & 8.76\%         \\ 
\hline
\textbf{DTRformer} & \textbf{14.50 } & \textbf{25.45} & \textbf{14.94\%} & \textbf{18.00} & \textbf{29.58} & \textbf{12.30\%} & \textbf{13.17} & \textbf{22.85} & \textbf{8.66\%} \\ 
\hline\hline
\end{tabular}
}
\label{table:ablation study}
\end{table}

- $w/o$ $E_a^{\text{imp}}$. It removes spatial-temporal adaptive embedding $E_a$, which represents the removal of the dynamic trend and uses only a combination of original inputs, temporal identity and predefined maps.

- $w/o$ $Transformer$. It removes Dynamic Spatial-Temporal Trend Transformer.

- $w/o$ $A^f$. It removes the forward pre-defined graph from the Multi-view Graph Fusion Generator Module.

- $w/o$ $A^b$. It removes the backward pre-defined graph from the Multi-view Graph Fusion Generator Module.

- $w/o$ $A^f \& A^b$. It removes the Multi-view Graph Fusion Module, which means that both the forward and backward graphs are removed.

- $w/o$ $augmented residual$. It removes the augmented residual from the AR-MSA module  of the connection.

\vspace{1em}
The results are shown in \cref{table:ablation study}. On the learning dynamic information aspect, $E^{\text{imp}}_a$, as an adaptive embedding, is indispensable for the learning of dynamic information.The Dynamic Spatial-Temporal Trend Transformer is important for modeling the inherent patterns in the traffic data. On the static information aspect, $A_f$ and $A_b$ describe the degree of influence of the forward and backward maps on the final prediction results, respectively, and the enhancement brought by the forward map on the PEMS08 dataset is less, but not negligible. After removing $A_f$ and $A_b$, the results show that our proposed Transformer can have more advanced results even when used directly for prediction. The results of applying augmented residual shows that for the fusion of multiple graphs of different types, adding more nonlinear information can help the model to converge better.
\subsection{Visualisation}

\begin{figure}[h]
\centering
\begin{subfigure}{0.5\textwidth}
        \centering
        \includegraphics[width=\linewidth]{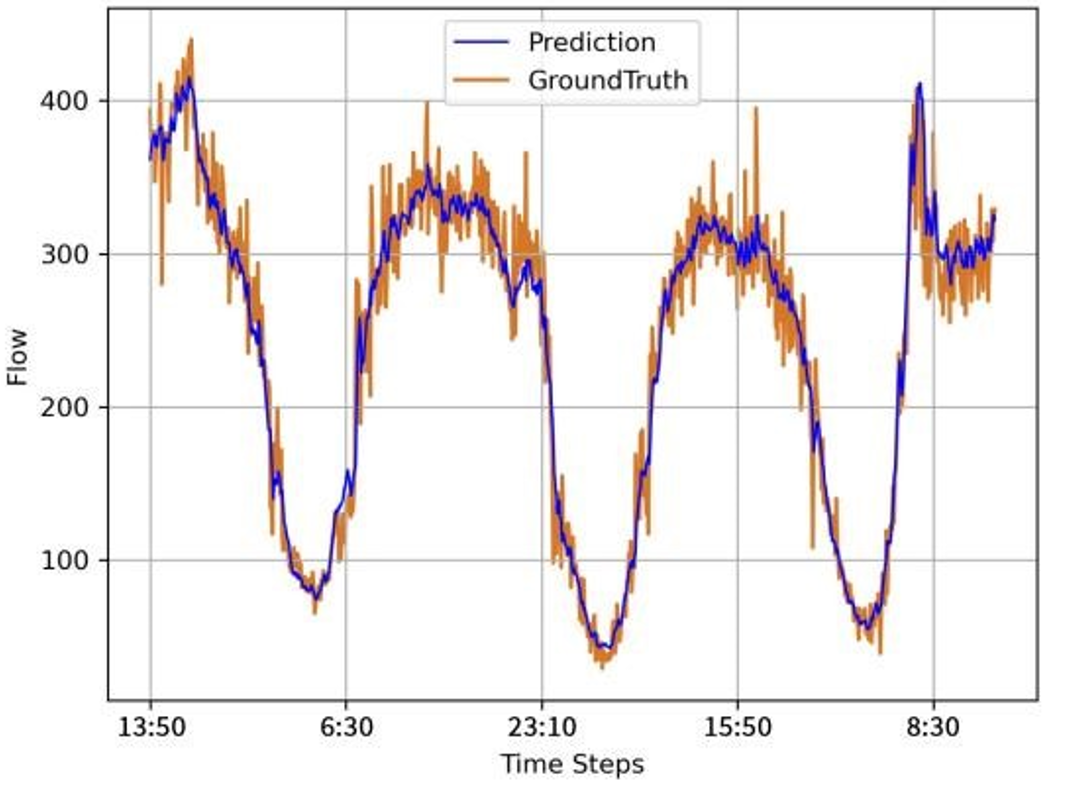}
        \caption{Visualization on node(sensor) 12}
        \label{fig:sub1}
    \end{subfigure}
    \hfill
\begin{subfigure}{0.5\textwidth}
        \centering
        \includegraphics[width=\linewidth]{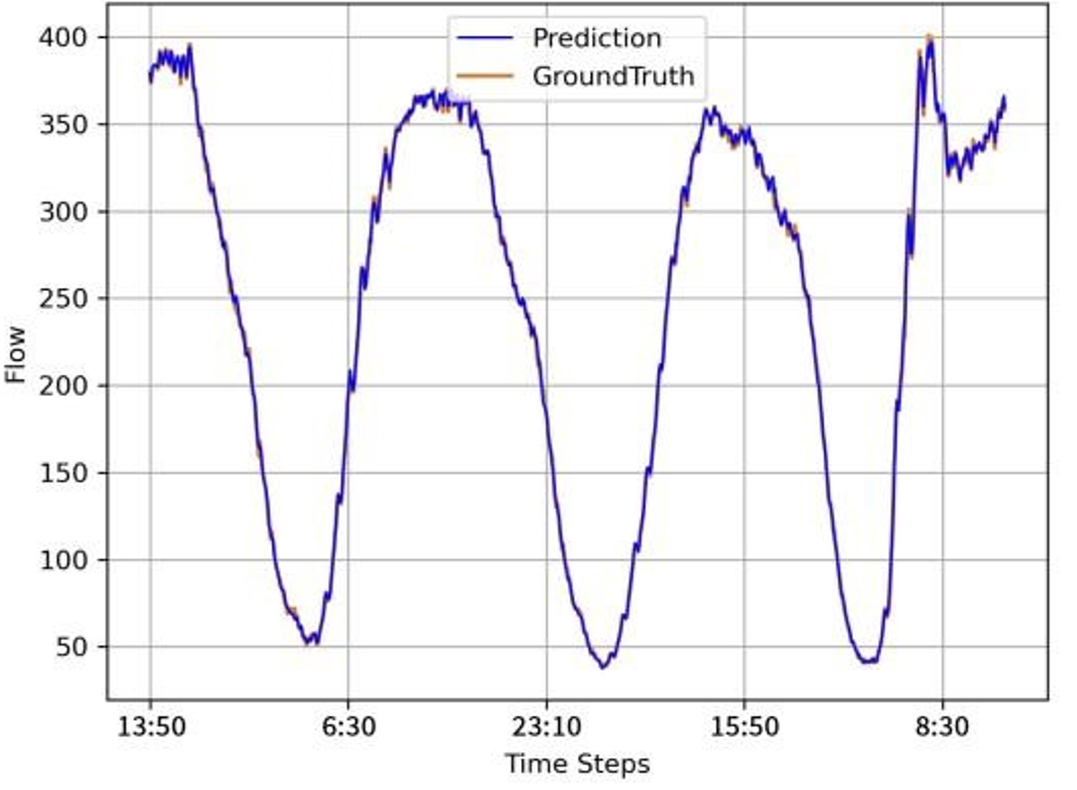}
        \caption{Visualization on node(sensor) 111}
        \label{fig:sub2}
    \end{subfigure}
\caption{Visualization of prediction results on PEMS08.}
\label{fig_ablation study}
\end{figure}
\subsubsection{Visualization of Predictions for Two Traffic Patterns}
In this section, the predictions of the model and the ground-truth data are visualized to further understand and evaluate the model. Two nodes are randomly selected, and three days of their data (from the test dataset) are presented. The prediction results for node 12 and node 111 are shown in \cref{fig_ablation study}. It is evident that the patterns of the two selected nodes are different. For instance, sensor 12 shows consistent congestion during the morning rush hour each day, enabling the model to fit the traffic variations reflected by sensor 12 very accurately. The figure shows minimal orange color, representing the ground-truth value, indicating high prediction accuracy for the more stable patterns. On the other hand, Sensor 111 indicates a road with consistently high traffic throughout the day, resulting in high values and long durations for most of the day. This implies amount of noise in the data. The results demonstrate that our model can capture these unique patterns at different nodes. Additionally, it is evident that the model effectively captures the inherent patterns of the time series while avoiding overfitting the noise. Overall, as depicted in \cref{fig_ablation study}, the model achieves impressive prediction accuracy. However, due to the presence of random noise, some local details may not be predicted accurately. For instance, the DTRformer model accurately capture the trend changes reflected by sensor 111 while successfully avoiding the overfitting of the large amount of noise signal.
\subsubsection{Visualization of Nodes Correlation}
\begin{figure}[h]
\centering
\includegraphics[width=0.45\textwidth]{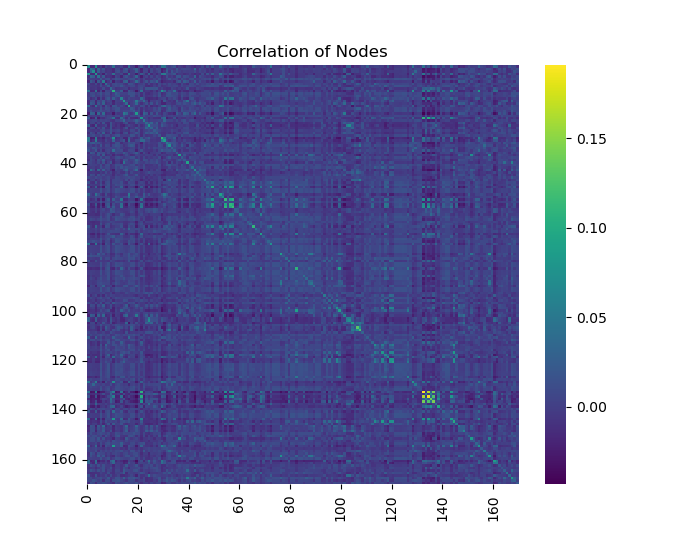}
\caption{Visualization of Nodes Correlation}
\label{Correlation}
\end{figure}

As shown in Fig.\ref{Correlation}, in the shown node correlation heatmap, brighter colors indicate that a node has a stronger influence on other nodes, as well as a greater degree of relationship to the final prediction results. We can observe that most nodes are not significant, with only a small number of key nodes making a substantial contribution to the overall prediction. Such results demonstrate that our linear projection method for the transportation network is effective.

\subsection{Efficiency}
In this part, the efficiency of DTRformer is compared with other methods based on the PEMS04 dataset. For a more intuitive and effective comparison, the average training time required for each epoch of these models is analyzed. All models are running on 13th Gen Intel(R) Core(TM) i7-13700KF @ 3.40 GHz, 64G RAM computing server, equipped with RTX 4090. The batch size is uniformly set to 16.

As shown in \cref{table3}, DGCRN \citep{Li2022} and D$^2$STGNN \citep{D2} are selected as representatives of dynamic graph and RNN combination structures, while GMAN \citep{Zheng2020} and PDFormer \citep{Jiang2023} are chosen as representatives of static graph and attention mechanism combinations. Both STAEformer \citep{Liu2023} and our proposed model utilize Identity Embedding and attention combinations, so they are included in the comparison. The results in the table demonstrate that DGCRN, which employs RNN-based dynamic graph computation methods, has the slowest inference speed, while D$^2$STGNN shows significant efficiency improvements because it does not compute the dynamic graph at every time step but rather at intervals of several time steps. DTRformer achieves a certain lead in inference speed compared to dynamic graph-based models. Among all attention mechanism-based models, DTRformer is only slower in inference speed than PDFormer. Compared to the STAEformer model, which uses the same combination module, DTRformer shows some improvement in efficiency while incorporating graph embedding. This indicates that our model exhibits competitive computational efficiency in the field of traffic prediction.

\begin{table}[htb]
\caption{Training time per epoch and inference time comparison. (Unit: seconds)}
\centering
\resizebox{0.5\textwidth}{!}{
\begin{tabular}{c|cc}
\hline\hline
Dataset     & \multicolumn{2}{c}{PEMS04}\\ \hline\hline
Model      & Training            & Inference\\ \hline\hline
DGCRN & 129.27 & 31.3\\ 
D$^2$STGNN & 75.94 & 15.72\\ \hline
GMAN & 401.26 & 31.072\\
PDFormer & 110.43 & 10.62\\ 
STAEformer & 91.10 & 15.88\\
\textbf{DTRformer} & 90.2 & 15.67\\
\hline\hline
\end{tabular}
}
\label{table3}
\end{table}

\section{Conclusion}
In this study, a novel model \textbf{DTRformer} is introduced for traffic prediction, comprising two key modules: the Dynamic Spatial-Temporal Trend Transformer (DST$^2$former) and the Multi-view Graph Fusion Module (MGFM). The DST$^2$former focuses on generating dynamic trends by capturing intricate spatio-temporal correlations, while the MGFM effectively fuses these dynamic trends with predefined graphs to enhance model performance.

Our extensive experiments on four real-world traffic datasets demonstrate that DTRformer significantly outperforms existing methods, underscoring its efficacy and robustness in traffic prediction tasks. Notably, even without the MGFM, the DST$^2$former alone achieves superior accuracy compared to traditional transformer methods, highlighting its standalone effectiveness. The inclusion of the MGFM further amplifies performance by reducing redundancy and integrating dynamic and static information.

These findings confirm that our proposed model not only advances the state-of-the-art but also offers a robust solution for capturing multi-view dynamic information in traffic networks.

\section*{Acknowledgements}
This work was supported by Fundamental Research Funds for the Provincial Universities of Zhejiang (No. GK239909299001-011) and  IOT Technology Application Transportation Industry R \& D Center Open Project Funding (No. 2023-05).

\hfill \break
\hfill \break
\hfill \break
\hfill \break
\hfill \break
\hfill \break
\hfill \break
\hfill \break
\hfill \break
\hfill \break
\hfill \break
\hfill \break
\hfill \break
\hfill \break
\hfill \break
\hfill \break
\hfill \break
\hfill \break
\hfill \break
\hfill \break
\hfill \break
\hfill \break
\hfill \break
\hfill \break


 \bibliographystyle{elsarticle-num-names} 
 \bibliography{sample}






\end{CJK*}

\end{document}